\documentclass[10pt,twocolumn,letterpaper]{article}

\usepackage{ulem}
\usepackage{comment}
\usepackage{pifont}
\usepackage{multirow}
\usepackage{placeins}

\usepackage[pagenumbers]{cvpr} %

\usepackage{xcolor}
\usepackage[table]{xcolor}

\newcommand{\bl}[1]{{\color{lightgray}#1}}
\newcommand{\de}[1]{{\footnotesize{\textcolor{blue}{(#1)}}}}
\newcommand{\debad}[1]{{\footnotesize{\textcolor{gray}{(#1)}}}}

\newcommand{\cmark}{\checkmark}

\renewcommand{\eg}{\textit{e.g.}\xspace}
\renewcommand{\ie}{\textit{i.e.}\xspace}

\usepackage{booktabs}
\usepackage{makecell}

\newcommand{\modelname}{AuditDM\xspace}

\definecolor{cvprblue}{rgb}{0.21,0.49,0.74}
\usepackage[pagebackref,breaklinks,colorlinks,allcolors=cvprblue]{hyperref}
\def\paperID{1465} %
\def\confName{CVPR}
\def\confYear{2026}

\title{Differences That Matter: \\ Auditing Models for Capability Gap Discovery and Rectification}

\author{
  Qihao Liu$^{1, 2}$\thanks{This work was done during Qihao Liu’s internship at Google.}\quad
  Chengzhi Mao$^{1}$\quad
  Yaojie Liu$^1$\quad
  Alan Yuille$^2$\quad
  Wen-Sheng Chu$^1$ \\
 {\normalsize $^1$Google \qquad $^2$Johns Hopkins University} \\
 {\small \texttt{\url{https://auditdm.github.io/}}}
}

\begin{document}
\maketitle

\begin{abstract}
Conventional evaluation methods for multimodal LLMs (MLLMs) lack interpretability and are often insufficient to fully disclose significant capability gaps across models.
To address this, we introduce {\bf \modelname}, an automated framework that actively discovers and rectifies MLLM failure modes by auditing their divergence.
\modelname fine-tunes an MLLM as an auditor via reinforcement learning to generate challenging questions and counterfactual images that maximize disagreement among target models.
Once trained, the auditor uncovers diverse, interpretable exemplars that reveal model weaknesses and serve as annotation-free data for rectification.
When applied to SoTA models like Gemma-3 and PaliGemma-2, \modelname discovers more than 20 distinct failure types.
Fine-tuning on these discoveries consistently improves all models across 16 benchmarks, and enables a 3B model to surpass its 28B counterpart.
Our results suggest that as data scaling hits diminishing returns, targeted model auditing offers an effective path to model diagnosis and improvement.
\end{abstract}

\vspace{-1ex}

\begin{figure}[t]
    \centering
    \captionsetup{type=figure}
    \vspace{-7mm}
    \includegraphics[width=\columnwidth]{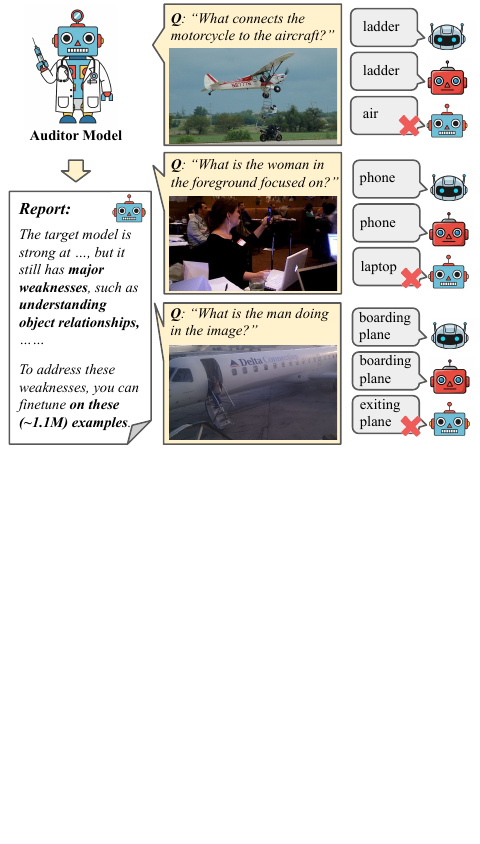} \vspace{-75mm}
    \captionof{figure}{\textbf{Overview of \modelname.}
    We propose to train an {\textit{auditor}} model to systematically discover capability gaps in an MLLM by generating failure-inducing question–image pairs.
    We show three automatically generated examples of weaknesses in object relationships.
    The proposed framework offers diagnostic insight and enables targeted rectification via auditor-guided feedback.
    }
    \vspace{-1ex}
    \label{fig:teaser}
\end{figure}%

\section{Introduction}

The rapid evolution of Multimodal Large Language Models (MLLMs) has led to a surge in high-performing models \cite{hurst2024gpt,wang2024qwen2,dubey2024llama,comanici2025gemini}.
However, despite steady improvements on public benchmarks, selecting the proper models for real-world deployment remains challenging, as conventional evaluations often obscure how models truly differ.
This is particularly evident when models are retrained on new data: although retraining may improve targeted knowledge, its impact on broader capabilities remains unclear. 
Similar concerns arise with task-specific fine-tuning and edge deployment, where practitioners must weigh trade-offs between accuracy, model size, and generalization.
In practice, the critical question is not \textit{``who wins the leaderboard''} but \textit{``what changes, and why''}: identifying which inputs flip, what skills improve, and where brittle behaviors persist.

While common benchmarks~\cite{liu2024mmbench,ying2024mmt,antol2015vqa,hudson2019gqa} are the de facto standard for model comparison, they fall short in two key respects for answering the above questions.
\textit{First}, closed-set evaluations are bounded by a fixed knowledge scope and inevitably leave blind spots.
As a result, comparisons based on closed sets can be inherently selective and biased.
\textit{Second}, benchmarks tend to compress complex behavior into sparse scores, thus obscuring heterogeneous shifts across data slices, whereas the most significant capability gaps are often entangled and concentrated in the long tail.
Prior work proposed human online testing~\cite{sheng2021human}, yet it is prohibitively expensive and time-consuming to scale.
To bridge this gap, we introduce \textbf{model auditing}, an automatic evaluation paradigm designed to uncover and interpret hidden divergence patterns missed by closed sets online.
An effective ``auditor'' must go beyond simple detection: \textit{it should systematically discover capability gaps, summarize interpretable weaknesses, and provide feedback to guide rectification and model improvement.}

To this end, we propose to \textbf{Audit} the \textbf{D}ifferences that \textbf{M}atter, and introduce \textbf{\modelname}, an annotation-free framework that exploits cross-model divergence to discover failure modes in a target MLLM (Fig.~\ref{fig:teaser}).
Within the context of VQA\footnote{We focus on VQA capability to study MLLM performance gaps.}, rather than relying on human inspection~\cite{sheng2021human}, \modelname trains an MLLM auditor that generates question–image exemplars maximizing response divergence among target models, thereby exposing capability gaps.
Concretely, \modelname learns to propose natural-language probes that (i) directly query the target MLLM (\eg, complex questions probing image details) or (ii) instruct an image diffusion model to synthesize targeted perturbations (\eg, counterfactual images).

We train the auditor with Group Relative Policy Optimization (GRPO)~\cite{shao2024deepseekmath}, iteratively refining it to generate instructions that maximize cross-model disagreement.
This RL-style approach enables optimization over an interpretable yet non-differentiable language interface, yielding human-interpretable failure patterns rather than isolated errors.
At inference, since the auditor has learned the patterns that induce failures of the target model, it can expose model blind spots in a single pass. 
This enables synthesizing large-scale, weakness-targeted data that can be used immediately for failure rectification and model improvement.

\begin{figure}[t]
  \centering
  \begin{subfigure}[t]{0.45\columnwidth}
    \includegraphics[width=\linewidth]{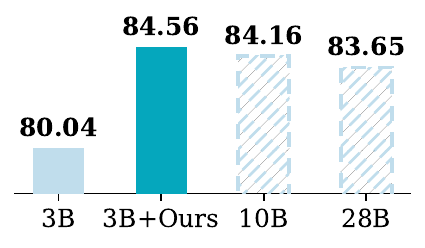}
    \caption{Improving PaliGemma2}
  \end{subfigure}
  \hfill
  \begin{subfigure}[t]{0.45\columnwidth}
    \includegraphics[width=\linewidth]{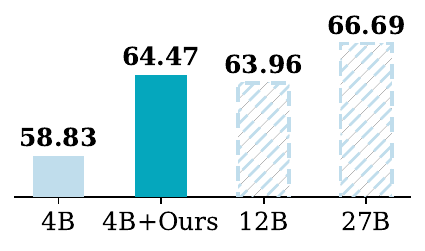}
    \caption{Improving Gemma3}
  \end{subfigure}
  \hfill

  \vspace{-2mm}
  \caption{\textbf{Model improvement with \modelname.}
    We report average performance over all benchmarks per model (excluding MME due to its incompatible score scale).
    Once trained, \modelname generates targeted, large-scale data points aligned with discovered weaknesses, training on which can produce consistent gains across diverse models and benchmarks.
    }
  \label{fig:teaser_perform}
  \vspace{-3mm}
\end{figure}

We validate \modelname through extensive experiments on Gemma3~\cite{team2025gemma} and PaliGemma2~\cite{beyer2024paligemma}.
Our analysis uncovers diverse failure modes within the PaliGemma2 family, directly pinpointing their capability boundaries and risk profiles.
Notably, \modelname identifies failure categories where a 28B model underperforms a 3B model, including hallucination avoidance, counting, and color recognition.
Moreover, we convert these diagnostic probes and counterfactuals into training signals, achieving substantial gains across 16 benchmarks.
For example, on AI2D~\cite{kembhavi2016diagram}, we raise PaliGemma2-3B from 76.0 to 85.3, even surpassing the PaliGemma2-28B model. 
On MMBench~\cite{liu2024mmbench}, we improve Gemma3-4B from 67.6 to 73.8, reaching performance comparable to its 12B counterpart.
As standard data sources near exhaustion, leveraging model disagreement for improvement provides a practical avenue for continual learning.
Our contributions are three-fold:
\begin{itemize}
    \item We introduce \textit{model auditing} as a new paradigm for systematically discovering model capability gaps and diagnosing weaknesses in MLLMs.
    \item We present \textbf{\modelname}, a reinforcement learning–based framework that trains an MLLM \textit{as an auditor} to automatically identify and interpret failure modes.
    \item We show that \modelname converts discovered failure modes into annotation-free training data, enabling targeted retraining and continual improvement.
\end{itemize}

\section{Related Work}

\begin{table*}[t]
\vspace{-3mm}
\caption{\textbf{Comparison of related work on finding and fixing MLLM failures}
}
\vspace{-2mm}
\label{tab:compar}
\centering
\small
\setlength{\tabcolsep}{5pt}
\resizebox{1\textwidth}{!}{
\begin{tabular}{ll|ccccccc}
{\bf Method} & & {\bf Data scale} & {\bf Weaknesses Seeking} & {\bf Image Weaknesses}  & {\bf Text Weaknesses} &  {\bf Failure Interpretability} &  {\bf Failure Rectification} \\
\midrule 
\multicolumn{2}{l|}{Conventional evaluation~\cite{antol2015vqa,kembhavi2016diagram, liu2024mmbench, ying2024mmt}} & fixed set & limited & - & - & - & - \\

\midrule 
\multirow{2}{*}{Attacks} & Visual adversarial attacks~\cite{goodfellow2014explaining} & open-ended & active & \cmark  & - & -  & adversarial only \\
                  & Jailbreak attacks~\cite{zou2023universal, zhang2025boosting, liu2023autodan} & open-ended & active & - & \cmark & - & - \\
                  
\midrule 
 & Caption generation~\cite{chen2024sharegpt4v} & fixed set & no & - & - & - & - \\
Data & Prompt rewriting~\cite{deng2025emerging} & open-ended & no& - & - & - & - \\
synthesis & Image synth/render~\cite{liu2024synthvlm} & open-ended & no & - & - & \cmark & - \\
& Concept perturbation~\cite{cascante2022simvqa, gupta2022swapmix} & fixed set & passive & limited & limited & \cmark & -  \\

\midrule 
\multicolumn{2}{l|}{{\bf \modelname} (ours)} & open-ended & active & \cmark & \cmark & \cmark & \cmark \\
\\
\end{tabular}}

\vspace{-5mm}
\end{table*}

\noindent\textbf{Multimodal Large Language Models.}
Recent advances have markedly improved the ability of LLMs to handle complex tasks~\cite{brown2020language, achiam2023gpt, touvron2023llama, bai2023qwen}, and have extended them with multimodal capabilities~\cite{baltruvsaitis2018multimodal, wang2024qwen2, hurst2024gpt, dubey2024llama, team2025gemma, beyer2024paligemma, comanici2025gemini}.
Early work like CLIP~\cite{radford2021learning} aligns visual and textual data within a shared representation space via contrastive learning~\cite{jia2021scaling, zhai2023sigmoid, sun2023eva, liu2025flowing}, establishing a common choice of visual encoder for MLLMs.
Recent approaches enhance multimodal integration by projecting visual outputs through Q‑Formers~\cite{bai2023qwenvl, li2023blip, dai2023instructblip, tong2024cambrian} or MLPs~\cite{mokady2021clipcap, liu2023visual, liu2024improved, zhu2023minigpt}, discretizing visual features to combine with text tokens~\cite{team2024chameleon}, or directly injecting continuous visual features into Transformer layers using Perceiver Resamplers~\cite{alayrac2022flamingo} or embedded visual expert  modules~\cite{wang2024cogvlm, zhang2023llama}.
Beyond architecture, training algorithms~\cite{ouyang2022training, dpo} and dataset curation~\cite{deng2025emerging} play a crucial role in driving the remarkable performance of modern MLLMs.
However, as data benefits plateau and evaluations lag, new strategies are needed to deepen understanding, strengthen evaluation, and drive further advances in MLLMs.
In this paper, we propose systematic auditing to uncover capability gaps and failure modes, evaluate inconsistencies, and guide targeted improvements.

\noindent\textbf{Adversarial attack for LLM/MLLM.}
Prior work on adversarial attacks mainly focuses on model safety, \eg, jailbreaks~\cite{zou2023universal, zhang2025boosting, liu2023autodan, ying2025jailbreak}, data exfiltration~\cite{greshake2023not, peng2024data,zhan2024injecagent}, and malicious intent~\cite{li2024images, liu2024discovering}, and largely relies on optimization-based methods~\cite{li2025exploiting, rando2024gradient}.
In contrast, our model auditing targets inherent weaknesses and recurring failure patterns without intentional attacks, proposing an optimization-free approach that identifies failures in a single inference step.

\noindent\textbf{Synthetic data for MLLM.}
MLLMs require extensive pretraining and fine-tuning data~\cite{radford2021learning, wang2024qwen2, liu2023visual}, but data collection is often labor-intensive and biased~\cite{paullada2021data}. 
To address this, synthetic generation is widely used: aligning images and text via GPT-4V-based captioning~\cite{chen2024sharegpt4v}, enriching text instructions with large pretrained models~\cite{deng2025emerging}, and rendering images from captions using diffusion~\cite{liu2024synthvlm}.
High-quality instruction-tuning sets are often constructed with formatted data~\cite{dai2023instructblip, chen2023x, zhang2023llama, wang2023visionllm, liu2023visual}, while diversity and complexity are increased through feature swapping or concept perturbations~\cite{cascante2022simvqa, gupta2022swapmix}.
In contrast, we move beyond generic alignment and diversity objectives by using an auditor to generate weakness-targeted samples guided by cross-model disagreement.
We generate data that directly closes capability gaps, improves evaluation fidelity, and enables label-free, continual rectification and model improvement.

\noindent\textbf{Self-Evolution and weak‑to‑strong learning.}
To advance LLMs toward, and potentially beyond, human-level performance, recent work explores self‑evolution methods (\eg, self‑play~\cite{tu2024towards}, self‑instruct~\cite{wang2022self}, self‑improving~\cite{huang2022large, zhuge2024agent}, self‑training~\cite{gulcehre2023reinforced}) and weak‑to‑strong learning~\cite{leike2023superalignment, burns2023weak}, enabling autonomous improvement under shifting environments.
Among them, self‑play unifies iterative improvements through adversarial dynamics induce curricula and self‑generated data with critics~\citep{chen2024self, yuan2024self, chen2025spc}. 
Extending to the zero‑data regime, recent work replaces external traces with task‑space exploration and language self‑play~\citep{zhao2025absolute, kuba2025language}.
While our approach draws on a similar principle of iterative improvement from self-generated data, we propose to explicitly train a model-specific \textit{auditor} for the target MLLM to uncover capability gaps and failure modes.
The auditor then synthesizes weakness-targeted data that closes these gaps and drives continual improvement.

\vspace{-0.4ex}
\section{\modelname}
\vspace{-0.3ex}
\begin{figure*}[t]
  \centering
  \vspace{-3mm}
  \includegraphics[width=0.99\linewidth]{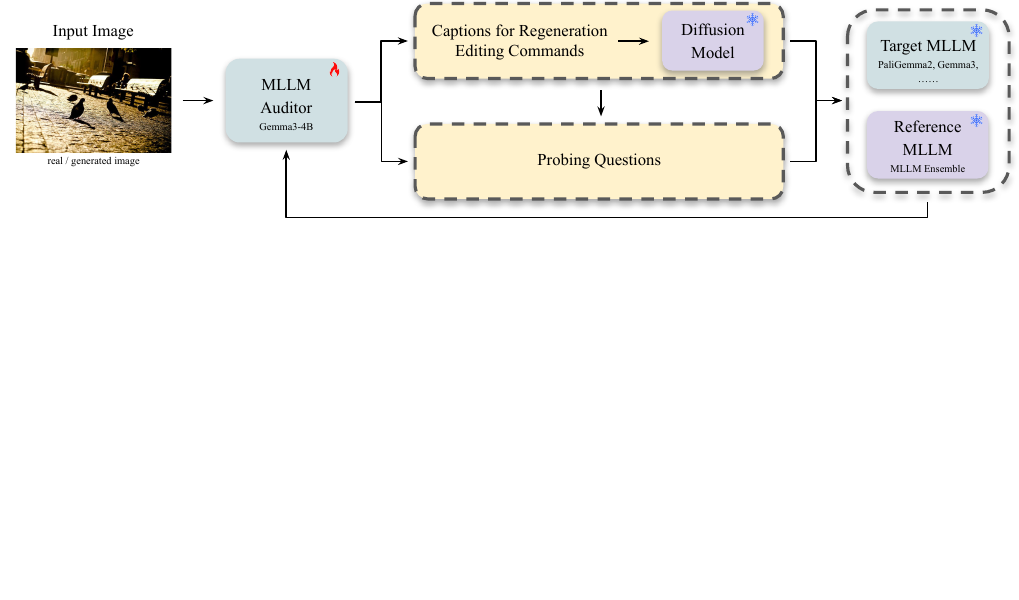}
  \vspace{-66mm}
  \caption{\textbf{\modelname architecture.}
  \modelname fine-tunes an MLLM into an \textit{auditor} that generates challenging probing questions and counterfactual images (via captions for image regeneration or editing commands), yielding question–image pairs on which the target model fails while the MLLM ensemble agrees, thus exposing capability gaps and failure modes.
  The auditor is trained to maximize prediction discrepancy between the target and the ensemble.
  Once trained, it identifies weaknesses and failure cases in a single inference pass.
  }
  \label{fig:pipeline}
  \vspace{-1ex}
\end{figure*}

\modelname is a reinforcement-learning (RL) framework for auditing MLLMs by actively discovering and rectifying their VQA failure modes.
As shown in Fig.~\ref{fig:pipeline}, we fine-tune an MLLM auditor and pair it with diffusion models to generate challenging question–image exemplars that maximize response disagreement between a target MLLM and an ensemble of reference MLLMs.
This section describes the process for generating failure-inducing pairs, ensembling reference MLLMs, training the RL-based auditor, and the mechanism for rectifying discovered failure modes.

\vspace{-0.1ex}
\subsection{Question-Image Pair Generation}
\label{sec:met:exemplar}

AuditLM first identifies question-images pairs that induce failures in a target MLLM.
Given an input image $I$ and prompt $p$, the auditor $\mathcal{A}$ analyzes the input image $I$ and generates textual outputs for two generation tasks, which together construct the failure-inducing pair $(Q^*,I^*)$.

\noindent\textbf{Text instruction for {\boldmath$I^*$} generation.} 
To probe visual weaknesses in the target MLLM, the auditor collaborates with \textit{either} an image diffusion model $\mathcal{G}$ or an image editing model $\mathcal{E}$ to create counterfactuals.
For \textit{image generation}, the auditor produces a detailed caption $C \!=\! \mathcal{A}(I,p_c)$ for the input image $I$, and enriches it with challenging semantic elements.
This caption is then used to condition the diffusion model to synthesize an altered image $I_g \!=\! \mathcal{G}(C)$ that embodies these challenging attributes.

For \textit{image editing}, which supports more targeted analysis, the auditor generates editing commands $E \!=\! \mathcal{A}(I,p_e)$ for an image-editing model to apply controlled modifications to produce the edited image $I_e \!=\! \mathcal{E}(I,E)$.
This process systematically tests how the target model responds to visual perturbations and thus reveals potential weaknesses.
While image editing yields fewer counterfactuals than full image synthesis, it offers more precise and interpretable insights into specific visual failure modes.

\noindent\textbf{Direct {\boldmath$Q^*$} generation.} 
The auditor also generates complex, nuanced questions $Q^*$ specifically designed to challenge the target MLLM with the instruction prompt for question generation $p_q$.
This step drives the auditor to identify intricate semantic concepts in the image, learn textual and visual patterns the target model struggles with, and craft targeted queries that probe these weaknesses. 
We generate $Q^*$ from either the original image $I$ or the counterfactual images ${I_g, I_e}$, \ie, $Q^* = \mathcal{A}(I',p_q), \text{where~} I' \in \{I, I_g, I_e\}$. 
In practice, we combine all three pairing levels, \ie, $(Q^*,I^*)$, $(Q^*,I)$, and $(Q,I^*)$. 
An ablation of their combinations is provided in Sec.~\ref{sec:exp:abl}.
The instruction prompts ${p_q, p_c, p_e}$ for all three generation tasks are provided in Sec~\ref{sec:supp:prompt}.

\subsection{Divergence of Models}
\label{sec:met:assumption}

Given the generated pairs $(Q^*,I^*)$, we then evaluate the target MLLM $\mathcal{M}_{tar}$ against a reference MLLM $\mathcal{M}_{ref}$.
When auditing the difference between two models, the second model is taken as the reference.
When auditing a single target for model weaknesses, we instead use an ensemble whose consensus serves as an oracle, and search for $(Q^*,I^*)$ that maximize divergence from this oracle.
By comparing responses from $\mathcal{M}_{tar}$ and $\mathcal{M}_{ref}$ , we quantify their disagreement to identify capability gaps and pinpoint failure modes.
In particular, inconsistencies between the target model's output and the ensemble's consensus provide an interpretable signal that exposes the target's deficiencies and risk areas.
This disagreement-driven evaluation is a critical diagnostic tool, offering a deeper understanding of the target MLLM's behavior and decision boundaries, which subsequently guides targeted updates to improve its performance (details in Sec.~\ref{sec:self_improve}).

To attribute observed failures to the target model $\mathcal{M}_{tar}$, rather than artifacts of the auditor (\eg, generating meaningless questions), diffusion model (\eg, generating inaccurate or unrealistic images), or ensemble (\eg, producing incorrect answers), we adopt two assumptions.
(1) \textit{Answerable instances:} If the ensemble agrees on an answer, the question–image pair is likely meaningful and answerable.
(2) \textit{Rarity of target correctness:} It is rare for $\mathcal{M}_{tar}$ to be uniquely correct while all ensemble models are wrong.
Under these assumptions, we treat the ensemble consensus as a strong proxy for correctness and update $\mathcal{A}$ only on instances where its prediction conflicts with that consensus.
These assumptions are empirically validated in Sec.~\ref{sec:supp:assumpt}.

\subsection{RL-based Auditor Training}
\label{sec:met:training}

We train the auditor $\mathcal{A}$ using Group Relative Policy Optimization (GRPO)~\cite{shao2024deepseekmath} to produce reliable $(Q^*, I^*)$ pairs that maximize discrepancies between the target model $\mathcal{M}_{tar}$ and the reference $\mathcal{M}_{ref}$. 
For each randomly generated pair $(Q^*,I^*)$ in Section~\ref{sec:met:exemplar}, we compute a disagreement signal $s(Q^*,I^*)$ by comparing the model responses:
\begin{align}
    s(Q^*,I^*) = D\Big(\mathcal{M}_{tar}(Q^*,I^*), \mathcal{M}_{ref}(Q^*,I^*)\Big),
\end{align}
where the distance metric $D$ is a binary semantic-agreement judge that returns 1 if the two answers differ in meaning and 0 otherwise. 
GRPO uses group-relative normalization of this signal to form advantages within each group:
\begin{align}
\hat{A}^{k}(Q^*,I^*) \;=\;
\frac{s^{k}(Q^*,I^*)-\mathrm{mean}_{j}\!\big[s^{j}(Q^*,I^*)\big]}
{\mathrm{std}_{j}\!\big[s^{j}(Q^*,I^*)\big]+\epsilon}.
\end{align}
We optimize the GRPO objective~\cite{shao2024deepseekmath} to train the auditor to favor examples that maximize cross-model discrepancies. 
By emphasizing these differences, the auditor learns to detect weaknesses and failure modes in $\mathcal{M}_{tar}$, produces valuable training signals, and ultimately serves as a fast, single-pass diagnostic tool for multimodal systems.

\subsection{Failure Mode Rectification}
\label{sec:self_improve}

Once trained, the target model's auditor is used to systematically identify failure modes and convert them into training signals.
However, fine-tuning on these failures is intuitive but difficult in practice~\cite{liu2023poseexaminer}.
We provide two main strategies to leverage these signals for rectifying such failure modes.

\noindent {\bf (1) Augmenting labeled data:} 
A standard strategy is to augment the original training data with auditor-generated samples (as detailed in Sec~\ref{sec:exp:self_imp:task}).
It creates a comprehensive training set that specifically covers identified weaknesses and failure cases.
This strategy mitigates overfitting to isolated examples, thereby improving the model's robustness and overall performance.

\noindent {\bf (2) Bootstrapping unlabeled data:}
To bootstrap data generation without labels, we leverage both the untrained auditor and auditors saved at different training steps to synthesize training samples (as in Sec.~\ref{sec:exp:self_imp:general}).
Specifically, for a collection of unlabeled images, each auditor produces a set of questions, new images, and corresponding pseudo-labels; we then aggregate and deduplicate them to form a combined training set for MLLM fine-tuning.
This process is iterated until performance converges: fine-tune the MLLM, refresh auditor checkpoints with the latest MLLM, and regenerate data from the unlabeled pool.

The auditor-synthesized, weakness-specific data is critical for refining the target MLLM and improving its real-world performance.
Starting from a base MLLM, we train an auditor to generate this weakness-specific data, and then fine-tune the MLLM on mixtures that include it.
We explore two data mixture strategies and leave broader strategy design to future work.
This iterative mechanism enables continual rectification and improvement of MLLMs: by leveraging evolving evaluations to refresh the training set, \modelname leads to higher benchmark accuracy, greater task robustness, and lower failure rates over time.

\section{Experiments}

We demonstrate the effectiveness of our proposed framework in evaluating, understanding, and improving MLLMs.
In Sec.~\ref{sec:exp:behavior}, we show that it efficiently and systematically uncovers weaknesses in a given MLLM, enabling clearer behavioral understanding and more faithful evaluation of its performance.
In Sec.~\ref{sec:exp:self-improve}, we show that it delivers significant performance gains across a range of benchmarks.

\noindent\textbf{Implementation details.} 
We evaluate \modelname on two target model families (PaliGemma2~\cite{steiner2024paligemma}, and Gemma3~\cite{team2025gemma}) by training auditors to automatically uncover their weaknesses and improve their performance, and we use a mixture of these model variants as the ensemble.
For PaliGemma2, we default to 448px$^2$ unless otherwise noted.
For all experiments, we fine‑tune Gemma3‑4B as the auditor, use FLUX.1‑dev~\cite{labs2025flux1} for image generation, and FLUX.1‑Kontext‑dev for image editing.
The auditor is fine-tuned for 1K steps with AdamW~\cite{loshchilov2017decoupled} using an initial learning rate of $3 \times 10^{-6}$ with a $10 \%$ warm-up and cosine learning rate decay to $1 \times 10^{-6}$, and a global batch size of 256.
Further details are provided in Sec.~\ref{sec:supp:imp_aud}.

\subsection{\modelname for Model Failure Detection}
\label{sec:exp:behavior}

In Sec.~\ref{sec:exp:eff}, we demonstrate the effectiveness of \modelname in finding model weaknesses; in Sec.~\ref{sec:exp:weak}, we provide a detailed analysis of the failure modes automatically discovered by our method.
We focus on PaliGemma2 here.

\subsubsection{Effectiveness of Finding Model Weaknesses}
\label{sec:exp:eff}

We demonstrate the effectiveness of the fine-tuned auditor in identifying model weaknesses.
The baseline uses the same system without fine-tuning, relying solely on prompt engineering to identify failure cases  (see Sec.~\ref{sec:supp:prompt} for the introduction prompts).
We randomly sample 20K image–question pairs from the VQAv2~\cite{antol2015vqa} training split
Then, for each pair, \modelname and the baseline each generate a new image–question pair to test the target model (PaliGemma2-3B~\cite{steiner2024paligemma}).
To ensure accurate evaluation, we use Gemini 2.5 Pro and the GPT-5 API, with human re-verification on disagreements, to produce reliable pseudo-annotations for the generated image–question pairs.
We then measure the \textit{search success rate}, defined as the fraction of generated samples that expose a validated error, \ie, the target model’s answer disagrees with the annotation.
As shown in Table~\ref{tab:baseline}, \modelname discovers failures far more efficiently than the baseline, achieving a substantially higher search success rate under the same generation budget.
Moreover, the weaknesses discovered by \modelname span a diverse set of skills (\eg, world knowledge, clock reading, size comparison; see Fig.~\ref{fig:model_weakness}) and are more explainable and interpretable (see Fig.~\ref{fig:model_edit}). 
Together, they provide a richer and more actionable understanding of the model’s weaknesses and failure modes.

\subsubsection{Weakness Analysis of PaliGemma2}
\label{sec:exp:weak}
\begin{table}[t]
\caption{\textbf{Effectiveness of finding model weaknesses.}
The baseline uses the same system without fine-tuning, relying solely on prompt engineering to find failures. 
We report the success rate of identifying valid errors over 20K attempts.
}
\vspace{-5mm}
\label{tab:baseline}
\begin{center}
\resizebox{0.8\columnwidth}{!}{
\begin{tabular}{l|cc}
& {\bf Baseline} & {\bf \modelname (Ours)} \\
\midrule
Search Success Rate & 21.4\% & 91.1\%  \\
\end{tabular}
}
\end{center}
\vspace{-7mm}
\end{table}

\begin{figure*}[t]
  \centering
  \vspace{-3mm}
  \includegraphics[width=\linewidth]{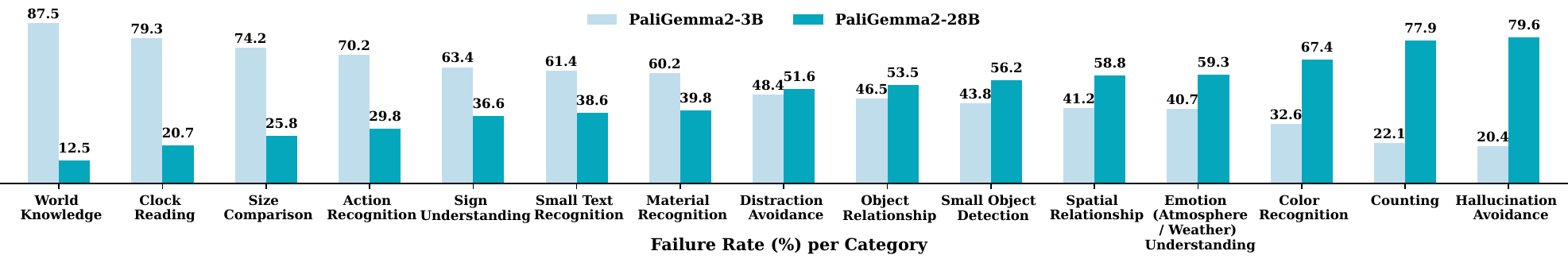} \vspace{-9mm}
  \caption{
  \textbf{\modelname identifies the top 15 failure modes and challenging task categories for PaliGemma2‑3B and 28B models at 448px$^2$, and we report normalized per-category failure rates.
  }
  Tasks are ordered left to right, beginning with the most pronounced weaknesses of the 3B model and progressing to those of the 28B.
  Notably, we observe that for certain tasks, the 28B model performs significantly worse than the 3B model.
  For example, on challenging images, the 28B model struggles more with color recognition and counting, and is more prone to hallucination.
  }
  \label{fig:model_weakness_dis}
\end{figure*}

\begin{figure*}[t]
  \centering
  \includegraphics[width=0.99\linewidth]{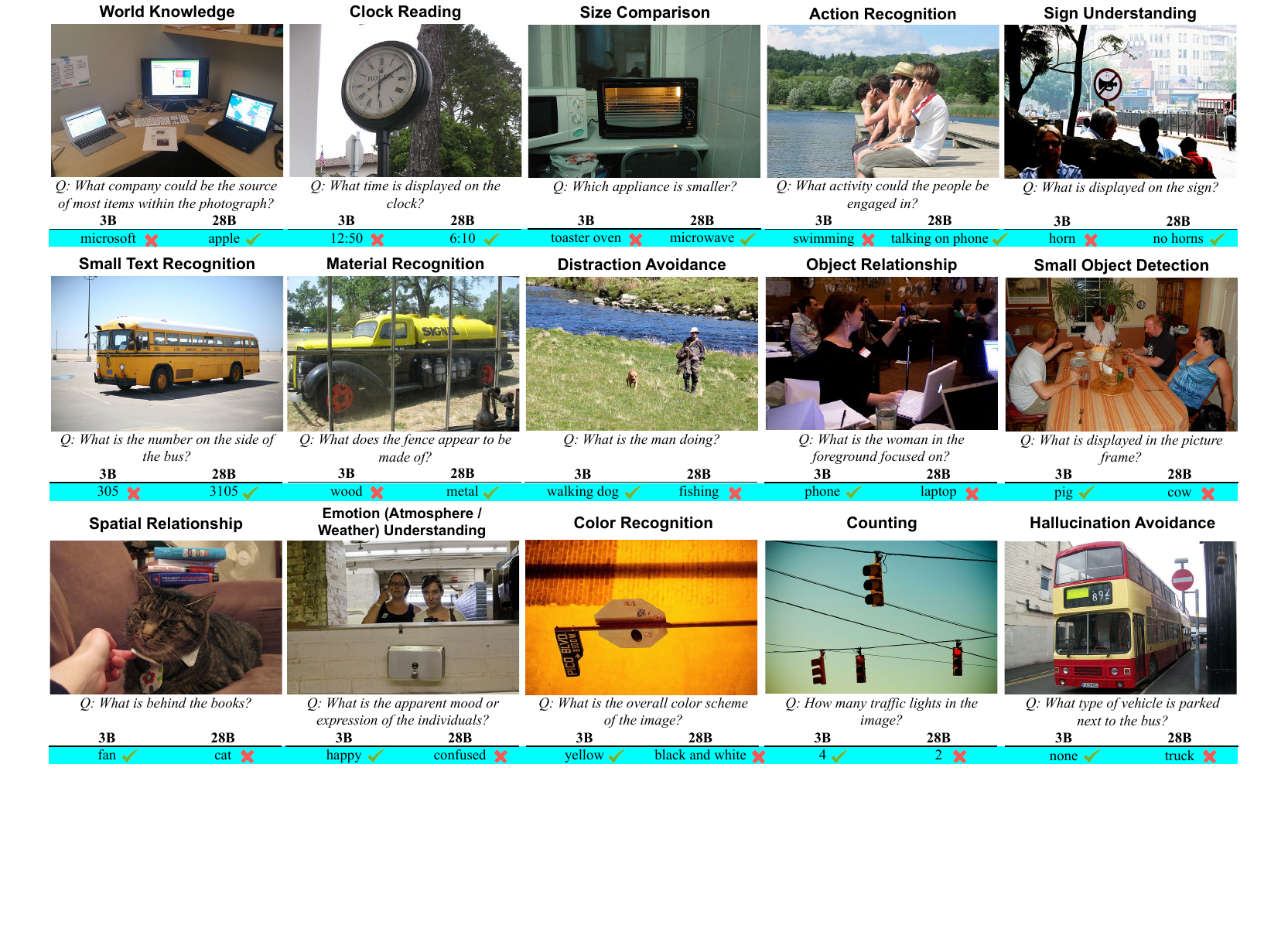} 
  \vspace{-23mm}
  \caption{
  \textbf{Generated examples for each failure category.}
  To better demonstrate the effectiveness, we focus on examples with \textit{original images} and \textit{generated questions}.
  Image-question pairs with both generated images and questions are provided in Fig.~\ref{fig:model_edit}.
  Some images are cropped or rotated for better figure layout. Original images and additional examples are provided in Sec.~\ref{sec:supp:qua}.
  }
  \label{fig:model_weakness}
\end{figure*}

\begin{figure*}[t]
  \centering
  \vspace{-1mm}
  \includegraphics[width=0.93\linewidth]{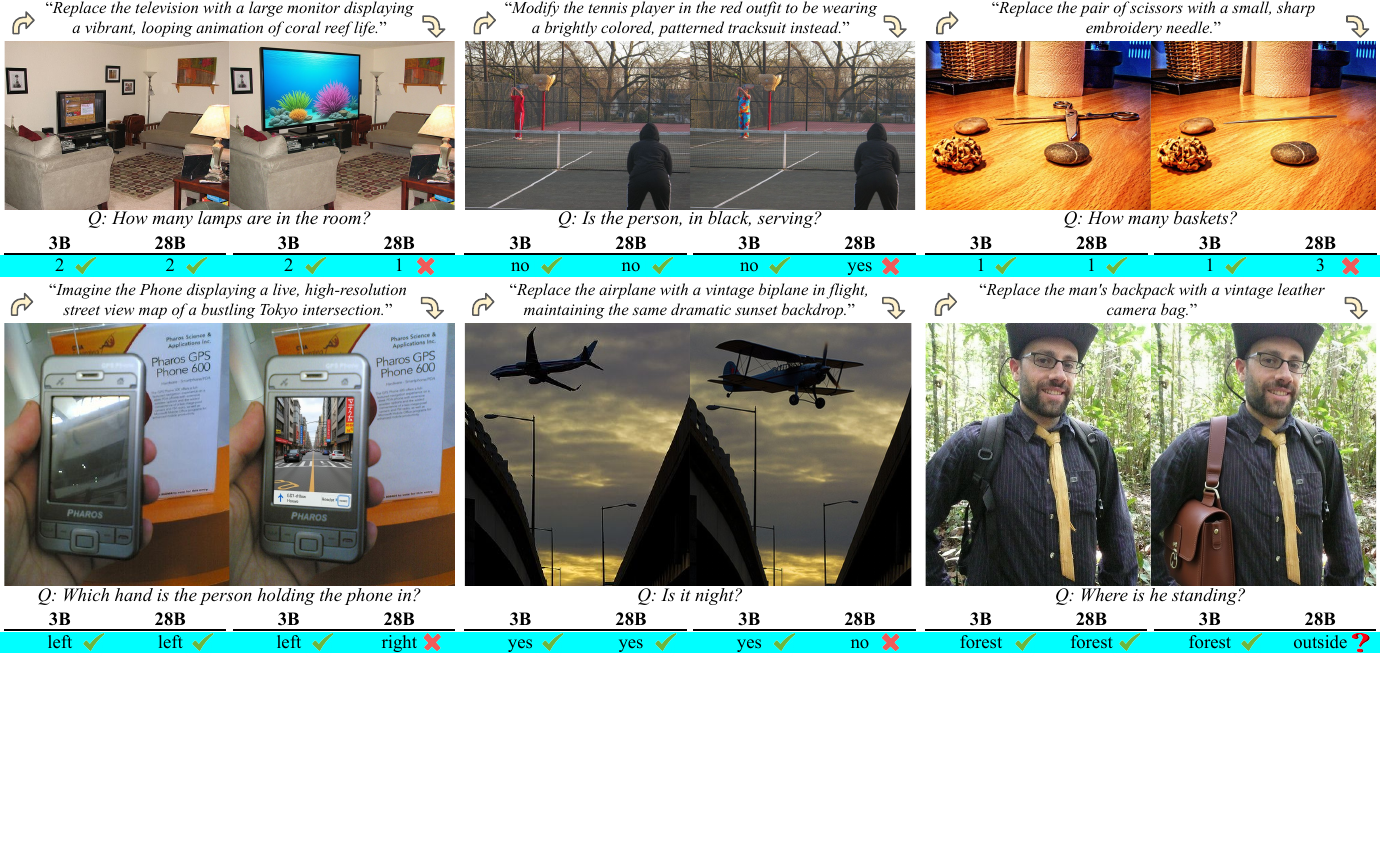} \vspace{-24mm}
  \caption{
  \textbf{\modelname efficiently detects fine-grained visual cues that are irrelevant to the task/question yet still alter model predictions.
  }
  We target PaliGemma2‑28B (448px$^2$) and showcase small modifications that fool the 28B model but not the 3B model, highlighting the effectiveness of our method in finding failures even in very powerful models.  
  For each example, we show the original image (left), the modified image (right), and the corresponding answers.
  Our analysis shows that even the PaliGemma-2 28B model is highly \textit{sensitive to minor changes in task-irrelevant objects}, suggesting that current MLLMs may still \textit{fail to ground visual reasoning in the correct evidence.}
  Our results further indicate that the 28B model \textit{is not necessarily more robust than the smaller 3B variant} and continues to exhibit various weaknesses.
  By isolating fine-grained cues that change predictions, \textit{\modelname reveals what visual evidence drives model behavior and affects outputs}, characterizing the model’s vulnerabilities and brittleness.  
  Additional examples are provided in Sec~\ref{sec:supp:qua}.
  }
  \label{fig:model_edit}
\end{figure*}

The PaliGemma2~\cite{steiner2024paligemma} family attains strong results across diverse benchmarks, with the 28B model significantly outperforming the 3B variant.
However, standard benchmarks report only aggregate performance on predefined tasks, whereas \modelname enables automatic discovery of weaknesses and failure modes.
In this section, we analyze their weaknesses and highlight cases where the 28B model underperforms the 3B model.

First, to ensure a fair comparison between the 3B and 28B models while removing potential confounds from image style, we consider \modelname in a setting that generates probing questions only for the original images.
Fig.~\ref{fig:model_weakness_dis} summarizes the top 15 failure patterns for the 3B and 28B models, automatically identified by \modelname, and Fig.~\ref{fig:model_weakness} provides representative examples for each failure mode.
For each detected failure, we use Gemma3 to summarize the root cause and the weakness category.
While \modelname identifies a wide range of limitations and challenges in current models, we present only a subset of the most notable weaknesses due to space constraints.
Our \modelname demonstrates that, compared with the 28B model, the 3B model shows pronounced weaknesses in world knowledge and struggles with tasks such as reading clocks, comparing sizes, recognizing actions, and understanding signs.
Interestingly, we also identify several weaknesses that are more prominent in the larger 28B model.
Specifically, the 28B model shows a significantly higher tendency toward hallucination errors, and struggles more with tasks related to emotion understanding, color recognition, and counting.
Additional details and examples are available in Sec~\ref{sec:supp:qua}.

In addition, by leveraging the image-editing component, \modelname efficiently discovers small visual modifications that substantially alter model predictions.
This capability clarifies which visual factors drive behavior and makes model weaknesses and vulnerabilities more interpretable (Fig.~\ref{fig:model_edit}).
Visualizations are provided in Fig.~\ref{fig:model_edit}.
Notably, we observe that certain minor alterations (which may be irrelevant to the task or question) can cause the 28B model to fail while the 3B model remains correct.
These observations suggest that the two models have different decision boundaries and rely on different cues for model predictions.
By systematically revealing prediction-altering elements, \modelname provides valuable interpretability that highlights the susceptibility and limitations of state-of-the-art models, showing how subtle perturbations can exploit unexpected sensitivities in large-scale model decision making.

\subsection{\modelname for Model Improvement}
\label{sec:exp:self-improve}
\vspace{-0.5ex}

\modelname also enables efficient, scalable synthesis of weakness-aligned training data.
Once the auditor is trained, a single inference on any image yields a tailored image–question pair.
In this section, we demonstrate effectiveness of \modelname for improving MLLMs.

\subsubsection{Per-task Fine-tuning Results}
\label{sec:exp:self_imp:task}

\begin{table*}[t]
\caption{\textbf{Per-task fine-tuning results.} 
Following PaliGemma2, each model is fine-tuned separately for each task.
\modelname yields substantial improvements to the 3B target model, allowing it to surpass the 10B ({\bf{bold}}) or 28B ({\setlength{\fboxsep}{1pt}\colorbox{yellow!25}{highlighted}}) variants on several benchmarks.
}
\vspace{-5mm}
\label{tab:genqa}
\begin{center}
\resizebox{1\textwidth}{!}{
\begin{tabular}{l|llllllll}
{\bf Model} & {\bf VQAv2} & {\bf GQA} & {\bf OK-VQA} & {\bf AI2D} & {\bf DocVQA} & {\bf ChartQA} & {\bf RefCOCO} & {\bf COCOCap} \\
\midrule 
\bl{PaliGemma2-10B 448px$^2$} & \bl{85.8} & \bl{68.3} & \bl{68.6} & \bl{84.4} & \bl{76.6} & \bl{66.4} & \bl{78.2} & \bl{145.0} \\
\bl{PaliGemma2-28B 448px$^2$} & \bl{85.8} & \bl{68.3} & \bl{70.6} & \bl{84.6} & \bl{76.1} & \bl{61.3} & \bl{77.3} & \bl{145.2} \\
\midrule
PaliGemma2-3B 448px$^2$ & 84.8 & 68.1 & 64.1 & 76.0 & 73.6 & 54.0 & 76.3 & 143.4 \\
PaliGemma2-3B 448px$^2$ + \modelname (Ours) & \cellcolor{yellow!25}{\bf86.7} \de{+1.9} & \cellcolor{yellow!25}{\bf 71.1} \de{+3.0} & {\bf 69.2} \de{+5.1} & \cellcolor{yellow!25}{\bf 85.3} \de{+9.3} & \cellcolor{yellow!25}{\bf 77.5} \de{+3.9} & \cellcolor{yellow!25}63.8 \de{+9.8} & \cellcolor{yellow!25}77.8 \de{+1.5} & {\bf 145.1} \de{+1.7}\\
\end{tabular}
}
\end{center}
\vspace{-3mm}
\end{table*}

\noindent\textbf{Experimental setup.}
Following PaliGemma2~\cite{steiner2024paligemma}, we evaluate across a range of academic benchmarks and fine-tune PaliGemma2-3B for each task.
We use \modelname to generate a synthetic dataset equal in size to the task’s training set (one new example per training instance), filter it, and mix it with the original data to form the final training set.
Our evaluation spans general VQA, text-oriented VQA, image segmentation, and image captioning, covering eight representative datasets.
Full details on benchmarks, task setup, metrics, data splits, and benchmark-specific training configurations are provided in Sec.~\ref{sec:supp:imp_pali}.

\noindent\textbf{Results.}
Table~\ref{tab:genqa} reports results.
Baselines are fine-tuned only on the original training set, whereas our method fine-tunes on a mixture of original data and auditor-generated examples. 
Our method consistently improves the target model across all benchmarks by a substantial margin.
Gains on grounding tasks that require precise bounding boxes are smaller, as synthesized or edited images can shift object locations and thus misalign bounding-box annotations (see Sec.~\ref{sec:exp:abl} for details). 
Notably, on several benchmarks, the 3B model fine-tuned with \modelname outperforms the official 28B model fine-tuned on the original dataset, highlighting the significant improvements enabled by \modelname.

\subsubsection{General Benchmark Results}
\label{sec:exp:self_imp:general}

\begin{table*}[t]
\caption{\textbf{General benchmark results.} 
By systematically identifying model failure modes, \modelname delivers significant gains across all tested benchmarks, and even enables the 4B model to outperform or match its 12B ({\bf{bold}}) or 27B ({\setlength{\fboxsep}{1pt}\colorbox{yellow!25}{highlighted}}) variant on several tasks.
}
\vspace{-5mm}
\label{tab:genbench}
\begin{center}
\resizebox{1\textwidth}{!}{
\begin{tabular}{l|llllllll}
{\bf Model} & {\bf MMBench-v1.1} & {\bf MMTBench} & {\bf Seed-Bench-IMG} & {\bf MME} & {\bf MMMU} & {\bf MMStar} & {\bf RealWorldQA} & {\bf POPE} \\
\midrule
\bl{Gemma3-12B} & \bl{73.8} & \bl{58.5} & \bl{70.6} & \bl{1517.3} & \bl{44.8} & \bl{55.7} & \bl{58.3} & \bl{86.0} \\
\bl{Gemma3-27B} & \bl{78.3} & \bl{59.2} & \bl{73.2} & \bl{1526.6} & \bl{49.7} & \bl{58.7} & \bl{62.5} & \bl{85.2}\\
\midrule
Gemma3-4B & 67.6 & 53.2 & 65.7 & 1376.0 & 39.6 & 46.1 & 54.5 & 85.1 \\
Gemma3-4B + \modelname (Ours) & {\bf 75.0} \de{+7.4} & {\bf 58.9} \de{+5.7} & {\bf 72.9} \de{+7.2} & 1450.3 \de{+74.3} & {\bf 45.2} \de{+5.6} & 52.4 \de{+6.3} & {\bf 61.4} \de{+6.9} & \cellcolor{yellow!25}85.5 \de{+0.4} \\
\end{tabular}
}
\end{center}
\vspace{-5mm}
\end{table*}

\noindent\textbf{Experimental setup.}
To better assess improvements under realistic user behavior, we further experiment with Gemma3-4B on eight widely used, task-diverse multimodal benchmarks, including MMBench~\cite{liu2024mmbench}, SEED-Bench~\cite{li2023seed}, and others.
These benchmarks provide comprehensive coverage across diverse subtasks and ability dimensions.
We follow recent work~\cite{li2024llava, wiedmann2025finevision} to collect images, apply additional filtering and cleaning, and then use \modelname to generate training samples.
For fair comparison, all models are evaluated with the same recipe using VLMEvalKit~\cite{duan2024vlmevalkit}.
Note that we rewrite the matching script to extract final answers from model outputs.
Further details on training data, benchmarks, and configurations are provided in Sec.~\ref{sec:supp:imp_gemma}.

\noindent\textbf{Results.} 
As shown in Table~\ref{tab:genbench}, \modelname delivers substantial gains over the target models on every benchmark without any human annotations, demonstrating strong model improvement.
Although larger base models (Gemma3-12B and Gemma3-27B) still lead on some tasks, augmenting Gemma3-4B with \modelname markedly narrows the gap and even surpasses the 12B model on Seed-Bench-IMG, MMMU, and RealWorldQA.
These results suggest that weakness-targeted auditing is a scalable, data-efficient path to improve multimodal models without additional supervision, positioning \modelname as an effective tool for uncovering and remedying capability gaps in MLLMs.

\subsubsection{Ablation on Different Auditing Components}
\label{sec:exp:abl}
\begin{table}[t]
\caption{\textbf{Ablation on different auditing components.} Results are reported for PaliGemma2‑3B at 224 px$^2$. A `-' indicates a performance drop as the amount of generated data increases. }
\vspace{-5mm}
\label{tab:abl_policy}
\begin{center}
\resizebox{0.82\linewidth}{!}{
\begin{tabular}{l|ccc}
& {\bf GQA} & {\bf RefCOCO} & {\bf AI2D} \\
\midrule
Baseline & 66.2 & 73.4 & 74.7 \\
\midrule
Probing question & 68.5 & - & 78.2 \\
Image generation & 66.9 & - & - \\
Image editing & 67.2 & 74.6 & 76.3 \\
\midrule
Best Combination & 69.8 & 74.6 & 79.4 \\
\end{tabular}
}
\end{center}
\vspace{-3mm}
\end{table}

\modelname comprises three auditing components that jointly learn textual and visual representations: (i) generating probing questions, (ii) synthesizing new images, and (iii) producing edited images.
Although our final system uses all three components, different combinations offer distinct benefits across tasks. 
We ablate their contributions in Table~\ref{tab:abl_policy}.

On general VQA tasks like GQA~\cite{hudson2019gqa}, all three policies yield performance gains.
Interestingly, image editing slightly outperforms image regeneration, despite regeneration uncovering more diverse weaknesses.
We hypothesize that image editing makes only small stylistic or bias-related changes, whereas regeneration can inject additional biases and artifacts from the generative model, thereby increasing distribution shift.
Notably, the probing‑question policy delivers the largest gains, indicating that \textbf{asking more informative, targeted questions is an especially effective way to improve MLLM performance.}

For dense prediction tasks like RefCOCO~\cite{refcoco}, carefully designed image editing yields consistent improvements. 
We filter generated editing commands to ensure that the target object’s position remains unchanged.
However, for these tasks, probing-question generation and image regeneration are more challenging, as it is difficult to produce accurate pseudo annotations for newly generated questions and images.
This limitation could be mitigated by using a stronger ensemble to provide more reliable annotations, or by generating questions for images that already have dense labels (\eg, SA-1B~\cite{sam}). We leave these directions to future work.

For text/diagram-oriented OCR tasks such as AI2D~\cite{kembhavi2016diagram}, regenerating entire images is challenging and degrades performance, likely because our diffusion model struggles to generate accurate diagrams.
By contrast, probing question is markedly more effective for this task than for the other two, suggesting that improving question quality is a more reliable way to strengthen diagram understanding.

\section{Limitations and Future Work}

Despite \modelname's effectiveness in finding failure modes and improving performance, it has two main limitations.

\noindent\textit{(1) Limitations in image generation}. 
As discussed in Sec.~\ref{sec:exp:abl}, \modelname requires images with dense annotations to generate probing questions for studying model behavior on tasks such as grounding and segmentation.
In addition, due to the difficulty of synthesizing images with dense text and complex diagrams, \modelname also struggles to regenerate images for text/diagram-oriented OCR tasks.
We believe this limitation can be mitigated by bootstrapping pseudo-labels from strong vision annotators and employing text/diagram-specialized generators.

\noindent\textit{(2) Computational complexity}. 
While a single inference pass can uncover a failure case, the pipeline relies on both an MLLM and a diffusion model, which makes large-scale dataset synthesis time-consuming.
For example, to fine-tune Gemma3-4B, we spend about 5 days on 8 H100 to generate new text–image pairs.
More runtime analysis is provided in Sec.~\ref{sec:supp:time}.
However, we emphasize that using LLMs or diffusion models for data generation is widely adopted but also time-consuming~\cite{chen2024sharegpt4v, liu2024synthvlm}.
Since all methods rely on a single inference pass to generate data, our overall time cost is comparable to theirs.

\section{Conclusion}

We introduce \modelname, an MLLM auditor that automatically diagnoses capability gaps and failure modes of a target model, enabling more faithful evaluation and a systematic understanding of its model weaknesses.
\modelname also enables large‑scale generation of training data tailored to failure modes and capability gaps, driving a closed loop of targeted data synthesis, retraining, and re‑auditing for continual improvement. 
Experiments on PaliGemma2 and Gemma3 show that \modelname effectively identifies failure modes and weaknesses in state-of-the-art models and delivers significant performance gains across diverse benchmarks, demonstrating both the effectiveness of our approach and the importance of auditing MLLM performance gaps.

{
    \small
    \bibliographystyle{ieeenat_fullname}
    \bibliography{main}
}

\clearpage
\setcounter{page}{1}
\maketitlesupplementary

\renewcommand{\thesection}{\Alph{section}}
\setcounter{section}{0}

\section*{Overview}

\noindent 
In the appendix, we provide additional information that could not be included in the main paper due to space limitations. 
The appendix is structured as follows:
\begin{itemize}[] %
\item Sec.~\ref{sec:supp:method}. Method and implementation details
    \begin{itemize}[]
    \item \ref{sec:supp:prompt}. Instruction prompts in \modelname
    \item \ref{sec:supp:imp_aud}. Auditor training
    \item \ref{sec:supp:imp_pali}. PaliGemma2 training
    \item \ref{sec:supp:imp_gemma}. Gemma3 training
    \end{itemize}
\item Sec.~\ref{sec:supp:exp}. Additional experimental results
    \begin{itemize}[]
    \item \ref{sec:supp:data}. Training Gemma3-4B on the same data without \modelname
    \item \ref{sec:supp:assumpt}. Empirical validation of assumptions
    \item \ref{sec:supp:time}. Computational complexity
    \end{itemize}
\item Sec.~\ref{sec:supp:qua}. Additional qualitative examples
\end{itemize}

\section{Method and Implementation Details}
\label{sec:supp:method}

\subsection{Instruction Prompts in \modelname}
\label{sec:supp:prompt}
As mentioned in Sec.~\ref{sec:met:exemplar}, the auditor $\mathcal{A}$ relies on three instruction prompts, $p_c$, $p_e$, and $p_q$, to guide the generation of image-regeneration captions $C = \mathcal{A}(I, p_c)$, image-editing commands $E = \mathcal{A}(I, p_e)$, and new questions $Q = \mathcal{A}(I', p_q)$.
We provide the prompts here:

\vspace{1ex}
\noindent \textbf{Prompt for image regeneration ($\mathbf{p_c}$):}
\vspace{-1ex}
\begin{verbatim}
You are given an image. Produce a 
detailed, literal caption that would 
allow a model to regenerate the image,
but also introduce small alterations to 
certain visual attributes. Return a 
single final caption describing the 
modified version only.
\end{verbatim}

\noindent \textbf{Prompt for image editing ($\mathbf{p_e}$):}
\vspace{-1ex}
\begin{verbatim}
You are given an image. Generate a 
single image-editing command that 
describes how to modify the image. 
The modification must remain plausible
in the real world. The command should 
be specific, actionable, and unambiguous.
Return the editing command only.
\end{verbatim}

\noindent \textbf{Prompt for question generation ($\mathbf{p_q}$):}
\vspace{-1ex}
\begin{verbatim}
You are given an image. Generate a 
single question that can be answered 
solely based on its visible content. 
Return the question only.
\end{verbatim}

In addition, in Table~\ref{tab:baseline}, we compare our model to a baseline to evaluate the failure search success rate.
The baseline uses the same system without fine-tuning, relying solely on prompt engineering to identify failure cases.
The prompts used for the baseline are provided below:

\vspace{1ex}
\noindent \textbf{Baseline prompt for image regeneration ($\mathbf{p^b_c}$):}
\vspace{-1ex}
\begin{verbatim}
You are given an image. Produce a 
detailed, literal caption that would 
allow a model to regenerate the image,
but introduce small changes that are
easy for models to get wrong. Return
a single caption describing the 
modified version only.
\end{verbatim}

\noindent \textbf{Baseline prompt for image editing ($\mathbf{p^b_e}$):}
\vspace{-1ex}
\begin{verbatim}
You are given an image. Generate a 
single image-editing command that 
makes a realistic change but is 
challenging for vision models. The 
modification must remain plausible
in the real world. The command 
should be specific, actionable, and 
unambiguous. Return the editing 
command only.
\end{verbatim}

\noindent \textbf{Baseline prompt for question generation ($\mathbf{p^b_q}$):}
\vspace{-1ex}
\begin{verbatim}
You are given an image. Generate a 
single question answerable from its
visible content but challenging for
vision-language models. Return the 
question only.
\end{verbatim}

\subsection{Auditor Training}
\label{sec:supp:imp_aud}
For all experiments in this paper, we fine-tune a pretrained Gemma3‑4B~\cite{team2025gemma} as the auditor, and use FLUX.1‑dev~\cite{labs2025flux1} for image generation, FLUX.1‑Kontext‑dev for image editing.
For the reference MLLM, we consider two scenarios:
(1) \textbf{Model comparison.} 
When training the auditor to compare the behavior of two models, we designate one as the target MLLM and the other as the reference MLLM, and fine-tune the auditor to generate image–question pairs that maximize the discrepancy between their outputs.
(2) \textbf{Single-model analysis.}
When training the auditor to probe the weaknesses and failure modes of a single target model, we take that model as the target MLLM, and construct a reference ensemble from PaliGemma2~\cite{steiner2024paligemma}, Gemma3~\cite{team2025gemma}, and Qwen2.5-VL~\cite{bai2025qwen2}. 
Note that the target model is not included in the ensemble.

During training, only the auditor is updated, while all other modules remain fixed.
The auditor is fine-tuned for $1,000$ steps with AdamW~\cite{loshchilov2017decoupled}, using an initial learning rate of $3 \times 10^{-6}$, a $10 \%$ warm-up, cosine learning rate decay to $1 \times 10^{-6}$, and a global batch size of 256.

\subsection{PaliGemma2 Training}
\label{sec:supp:imp_pali}
We provide additional experimental details for PaliGemma2 here. For all experiments involves PaliGemma2, we default to $448$px$^2$ unless otherwise noted.

\noindent\textbf{Benchmarks.}
Following PaliGemma2~\cite{steiner2024paligemma}, we evaluate on a range of academic benchmarks and fine-tune PaliGemma2-3B separately for each task.
We consider four common tasks: general VQA, text-oriented VQA, image segmentation, and image captioning.
For general VQA, we use VQAv2~\cite{antol2015vqa}, GQA~\cite{hudson2019gqa}, and OK-VQA~\cite{marino2019ok}; for text-oriented VQA, we use AI2D~\cite{kembhavi2016diagram}, DocVQA~\cite{mathew2021docvqa}, and ChartQA~\cite{masry2022chartqa}; for segmentation and captioning, we use RefCOCO~\cite{referitgame} and COCOCap~\cite{lin2014microsoft}, respectively.
Following PaliGemma~\cite{beyer2024paligemma}, the models are trained on the corresponding training splits and evaluated on the validation or test splits, depending on their availability.
For the VQA tasks, we report exact-match accuracy; for RefCOCO, we report mean Intersection-over-Union (mIoU); and for COCOCap, we report CIDEr score.

\begin{table}[t]
\caption{\textbf{PaliGemma2-3B training hyperparameters for each task.}}
\vspace{-5mm}
\label{tab:hyper}
\begin{center}
\resizebox{\linewidth}{!}{
\begin{tabular}{l|cccccc}
& {\bf Epochs} & {\bf \makecell{Batch\\size}} & {\bf \makecell{Learning \\ rate}} & {\bf \makecell{ Weight \\ Decay}} & {\bf \makecell{LLM \\ dropout}} & {\bf \makecell{Label \\ smoothing}} \\
\midrule
VQAv2 & 14 & 256 & $1\times 10^{-5}$ & $1\times 10^{-7}$ & 0.0 & 0.0 \\
GQA & 4 & 256 & $1\times 10^{-5}$ & $1\times 10^{-7}$ & 0.0 & 0.0 \\
OK-VQA & 10 & 256 & $5\times 10^{-6}$ & 0.0 & 0.0 & 0.0 \\
AI2D & 16 & 256 & $1\times 10^{-5}$ & $1\times 10^{-6}$ & 0.0 & 0.0 \\
DocVQA & 10 & 256 & $1\times 10^{-5}$ & $1\times 10^{-6}$ & 0.0 & 0.0 \\
ChartQA & 40 & 256 & $1\times 10^{-5}$ & $1\times 10^{-6}$ & 0.1 & 0.2 \\
RefCOCO & 120 & 256 & $1\times 10^{-5}$ & 0.0 & 0.0 & 0.3 \\
COCOCap & 8 & 256 & $1\times 10^{-5}$ & $1\times 10^{-6}$ & 0.0 & 0.0 \\
\end{tabular}
}
\end{center}
\vspace{-3mm}
\end{table}

\noindent\textbf{Training configurations.}
For each task, we first fine-tune an auditor on a random subset of the training data using the hyperparameters in Sec.~\ref{sec:supp:imp_aud}.
Specifically, since each model is trained with a global batch size of 256 for $1,000$ steps, the auditor is exposed to at most 256K images during fine-tuning.
To construct the training pool for the auditor, we first filter the training set by image resolution and then randomly select 256K images to form the image pool for auditor training.

After training the auditor, we use it to generate new image–question pairs from the training data, producing one new pair for each original example.
We then mix these generated pairs with the original training set to further fine-tune the PaliGemma2-3B model at 448px$^2$ resolution.
Following PaliGemma~\cite{beyer2024paligemma}, all models are trained until convergence. We provide hyperparameters for each task in Table~\ref{tab:hyper}

\subsection{Gemma3 Training}
\label{sec:supp:imp_gemma}

To further demonstrate the effectiveness of \modelname for model failure rectification and model improvement, we also apply \modelname to Gemma3-4B and evaluate it on eight widely used, task-diverse multimodal benchmarks. We provide the experimental details below.

\noindent\textbf{Training data.}
\modelname requires only images as input. 
Therefore, we collect data from DataComp-1B~\cite{gadre2023datacomp} and a mixture of academically annotated datasets, following LLaVA-OneVision~\cite{li2024llava} and FineVision~\cite{wiedmann2025finevision}.
In addition to standard image filters such as resolution and caption alignment, we also aim to ensure sufficient diversity and complexity in the images. 
To this end, we follow DenseWorld~\cite{li2025denseworld} and generate dense understandings of all filtered images, which allows us to further select images based on factors such as the number of objects, the diversity of object categories and relationships, overall scene richness, and the presence of fine-grained visual details.
This additional filtering step removes overly simple, overly complex, or low-content images and ensures that the final training set contains images that can give rise to challenging yet meaningful probing questions or counterfactual images.
As a result, we obtain a pool of 1.3M images for training, including 736K from DataComp-1B and 600K from the remaining academically annotated datasets.

\noindent\textbf{Benchmarks.}
We use the VLMEvalKit framework~\cite{duan2024vlmevalkit} to evaluate all models on a diverse suite of 8 benchmarks, including MMBench-v1.1~\cite{liu2024mmbench}, MMTBench~\cite{ying2024mmt}, Seed-Bench-IMG~\cite{li2023seed}, MME~\cite{fu2023mme}, MMMU~\cite{yue2024mmmu}, MMStar~\cite{chen2024we}, RealWorldQA~\cite{realworldqa2024}, and POPE~\cite{li2023evaluating}.

\noindent\textbf{Training details.}
After constructing the training image set, we train the auditor for 1,000 steps on these images to identify weaknesses of the Gemma3-4B model, saving checkpoints every 250 steps.
The resulting auditors (checkpoints saved at different steps) are then used to generate new image–question pairs. 
We aggregate and deduplicate these images, and then use them to further fine-tune the Gemma3-4B model.
We subsequently train new auditors on the newly fine-tuned Gemma3-4B model to analyze its remaining weaknesses.
This sequence, consisting of (i) training the auditor with the latest MLLM, (ii) regenerating data from the unlabeled pool using the trained auditor, and (iii) fine-tuning the MLLM with the new data, constitutes one refinement iteration, and we perform two such iterations.
While we observe further improvements with additional iterations in preliminary experiments, we restrict ourselves to two iterations to keep the compute cost within a reasonable budget (see Sec.~\ref{sec:supp:time}).
In each iteration, the auditor is trained following the hyperparameters in Sec. \ref{sec:supp:imp_aud}.
Then, given the newly generated samples, we fine-tune Gemma3-4B with a global batch size of 512 and a learning rate of $2 \times 10^{-6}$ for one epoch.

\begin{table*}[t!]
\caption{\textbf{Training Gemma3-4B on the same images without \modelname.} 
}
\vspace{-5mm}
\label{supp:tab:genbench}
\begin{center}
\resizebox{1\textwidth}{!}{
\begin{tabular}{l|llllllll}
{\bf Model} & {\bf MMBench-v1.1} & {\bf MMTBench} & {\bf Seed-Bench-IMG} & {\bf MME} & {\bf MMMU} & {\bf MMStar} & {\bf RealWorldQA} & {\bf POPE} \\
\midrule
Gemma3-4B & 67.6 & 53.2 & 65.7 & 1376.0 & 39.6 & 46.1 & 54.5 & 85.1 \\
Gemma3-4B (with the same images) & 69.6 \debad{+2.0} & 54.8 \debad{+1.6} & 66.9 \debad{+1.2} & 1321.6 \debad{-54.4} & 40.5 \debad{+0.9} & 45.6 \debad{-0.5} & 56.3 \debad{+1.8} & 84.2 \debad{-0.9} \\
Gemma3-4B + \modelname (Ours) & 75.0 \de{+7.4} & 58.9 \de{+5.7} & 72.9 \de{+7.2} & 1450.3 \de{+74.3} & 45.2 \de{+5.6} & 52.4 \de{+6.3} & 61.4 \de{+6.9} & 85.5 \de{+0.4} \\
\end{tabular}
}
\end{center}
\vspace{-5mm}
\end{table*}

\section{Additional Experimental Results}
\label{sec:supp:exp}

\subsection{Training Gemma3-4B without \modelname}
\label{sec:supp:data}

In the main paper, we demonstrate that \modelname significantly improves the performance of Gemma3-4B on various benchmarks. 
To further verify that these gains come from \modelname rather than from additional fine-tuning on the selected images, we also fine-tune Gemma3-4B on the same images under the same settings. 
Since some of the images do not have original questions, we use a pre-trained Gemma3-4B model (the same initial model used by \modelname) to generate new questions. 
The results are shown in Table~\ref{supp:tab:genbench}. 
We observe that fine-tuning on the same data without \modelname yields only marginal improvements on a few benchmarks, whereas our method achieves substantial gains across all benchmarks.

\subsection{Empirical Validation of Assumptions}
\label{sec:supp:assumpt}
\begin{table}[t]
\caption{\textbf{Empirical validation of assumptions.}}
\vspace{-5mm}
\label{tab:assumption}
\begin{center}
\resizebox{0.7\linewidth}{!}{
\begin{tabular}{l|c}
{\bf Type} & {\bf Percentage (\%)} \\
\midrule
Target failures & 81.3 \\
Ambiguous questions & 11.5 \\
Unanswerable questions & 7.2 \\
\end{tabular}
}
\end{center}
\vspace{-3mm}
\end{table}

As discussed earlier, to ensure that the observed failures are attributable to the target model rather than artifacts of the auditor (e.g., generating meaningless questions), the diffusion model (e.g., producing inaccurate or unrealistic images), or the ensemble (e.g., providing incorrect answers), we rely on two assumptions:
(1) \textit{Answerable instances:} when the ensemble models agree on an answer, the question–image pair is likely to be meaningful and answerable.
(2) \textit{Rarity of target correctness:} it is relatively rare for target model to be uniquely correct while all models in the ensemble are wrong.

In this section, we show that the ensemble indeed identifies genuine failures of the target model, i.e., (1) the question–image pairs are meaningful and answerable, and (2) they expose weaknesses of the target model.
Specifically, we use the auditor to generate 1,000 new question–image pairs on which the target model and the ensemble disagree. 
We then manually verify the correctness of their answers. 
We report (i) the percentage of samples that truly expose weaknesses of the target model (\ie, \textit{target failures}), (ii) the percentage of samples where the target model is still accurate despite disagreeing with the ensemble (\ie, \textit{ambiguous questions}, where both answers can be considered correct), and (iii) the percentage of samples where the question itself is not answerable from the image (\ie, \textit{unanswerable questions}).
The results are provided in Table~\ref{tab:assumption}. 
We can see that 81.3\% of the samples indeed expose genuine weaknesses of the target model. 
In addition, we expect a higher percentage of target failures when using a more powerful ensemble.

\subsection{Computational Complexity}
\label{sec:supp:time}
To improve a target model such as Gemma3-4B, \modelname consists of three steps: (1) training an auditor model, (2) generating large-scale data with the trained auditor, and (3) fine-tuning the target model.
In our experiments with Gemma3-4B, the first stage requires 29 hours on 16 H100 GPUs for 1,000 training steps, and the second stage requires 63 hours on 16 H100 GPUs to generate all images using Ray implementation.
However, we emphasize that using LLMs or diffusion models for data generation is already a widely adopted but time-consuming practice~\cite{chen2024sharegpt4v, liu2024synthvlm}. 
As all these methods depend on inference with an LLM or diffusion model for data generation, our overall time cost is comparable to theirs.

\section{Additional Qualitative Examples}
\label{sec:supp:qua}

\begin{figure*}
  \centering
  \includegraphics[width=0.99\linewidth]{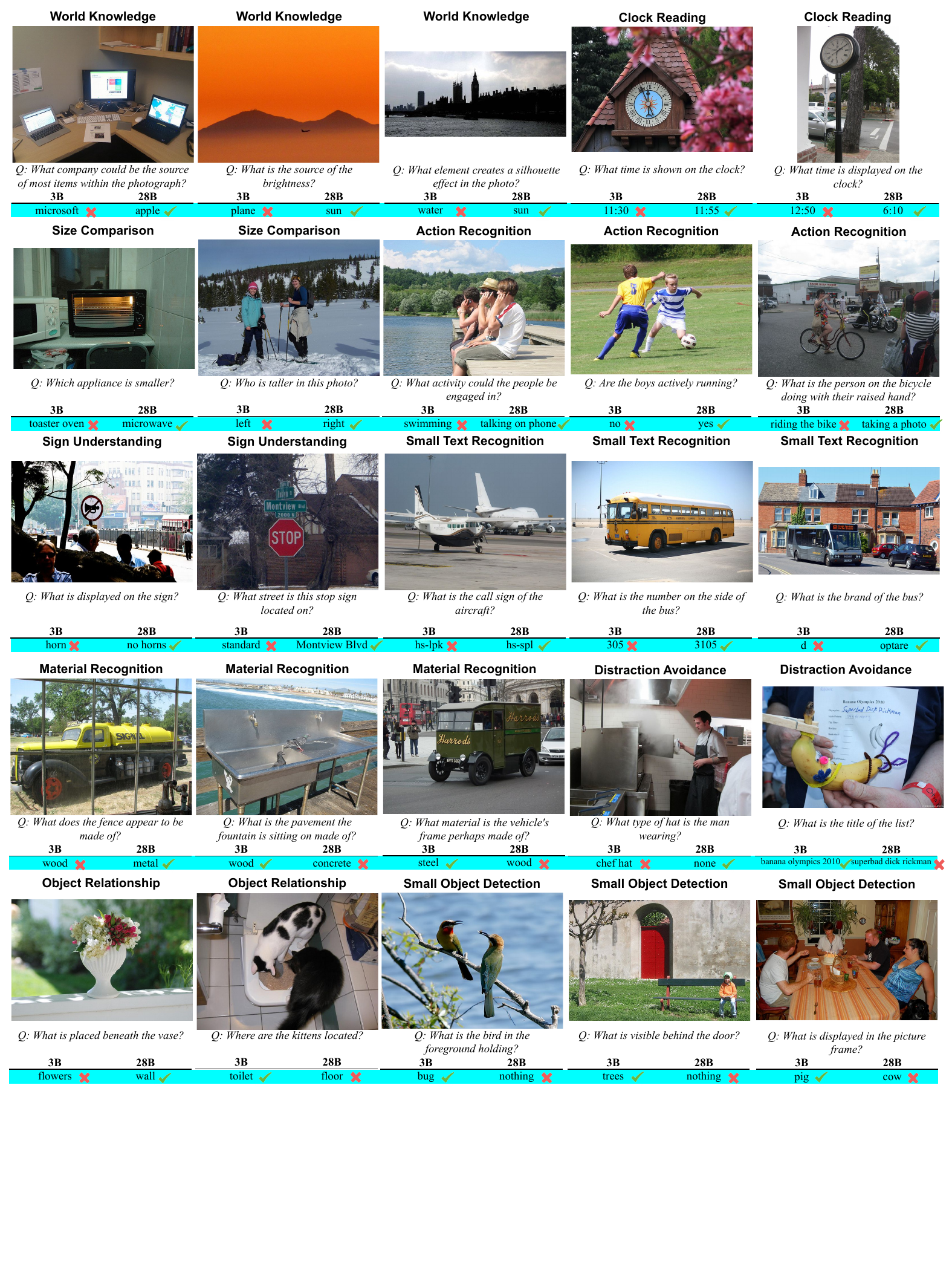} 
  \vspace{-33mm}
  \caption{
  \textbf{Generated examples for each failure category.}
  To better demonstrate the effectiveness, we focus on examples with \textit{original images} and \textit{generated questions}.
  }
  \label{fig:model_weakness_supp1}
\end{figure*}

\begin{figure*}
  \centering
  \vspace{-2mm}
  \includegraphics[width=0.99\linewidth]{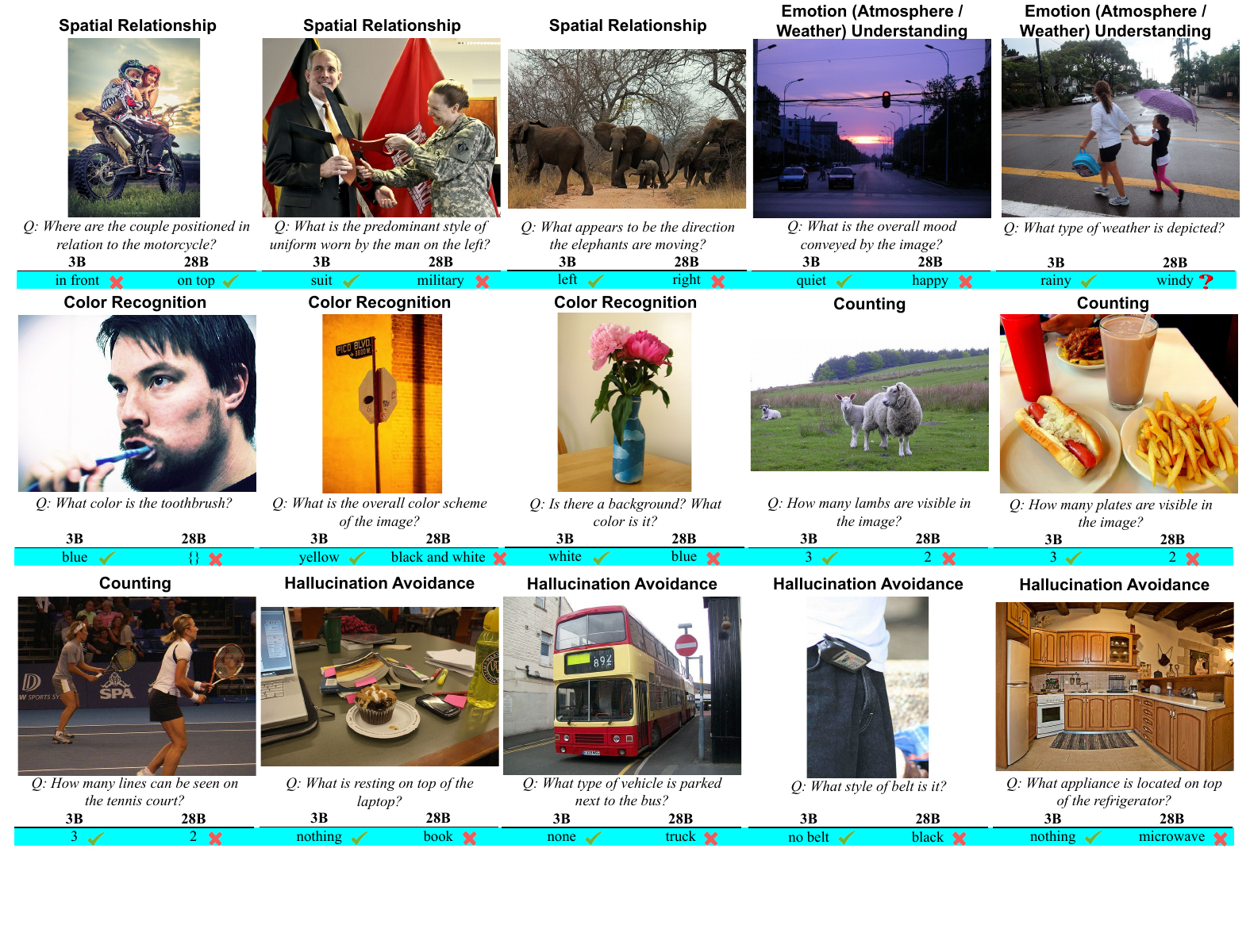} 
  \vspace{-17mm}
  \caption{
  \textbf{Generated examples for each failure category.}
  To better demonstrate the effectiveness, we focus on examples with \textit{original images} and \textit{generated questions}.
  }
  \label{fig:model_weakness_supp2}
\end{figure*}

\begin{figure*}[t]
  \centering
  \includegraphics[width=0.93\linewidth]{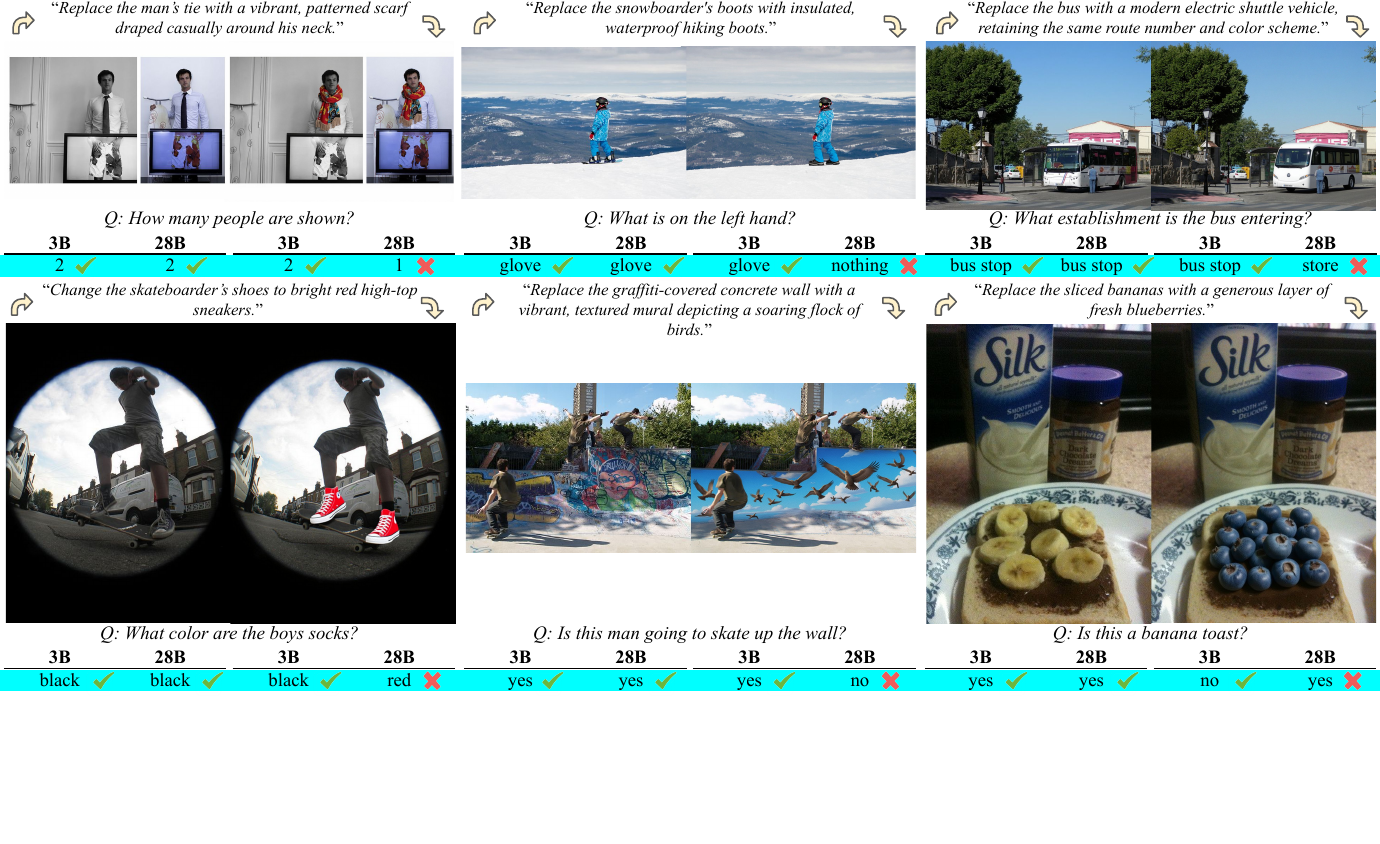} \vspace{-20mm}
  \caption{
  \textbf{Additional examples of fine-grained visual cues that are irrelevant to the task or question but still alter the model’s predictions.
  }
  We target PaliGemma2‑28B (448px$^2$) and showcase small modifications that fool the 28B model but not the 3B model, highlighting the effectiveness of our method in finding failures even in very powerful models.  
  For each example, we show the original image (left), the modified image (right), and the corresponding answers.
  }
  \label{fig:model_edit_supp}
\end{figure*}

We provide additional examples of the failures identified by our model in PaliGemma2 in Fig.~\ref{fig:model_weakness_supp1}, Fig.~\ref{fig:model_weakness_supp2}, and Fig.~\ref{fig:model_edit_supp}. 

Specifically, Fig.~\ref{fig:model_weakness_supp1} and Fig.~\ref{fig:model_weakness_supp2} show additional examples in their original image ratios for each weakness category. 
Note that, to ensure a fair comparison between the 3B and 28B models while eliminating potential influences of image style, we generate probing questions on the original images without altering their visual appearance. 
These results further confirm that the identified failures stem from intrinsic model limitations rather than superficial visual differences.

Fig.~\ref{fig:model_edit_supp} presents additional examples involving visual changes. 
Our model efficiently discovers small image modifications that significantly alter the model’s predictions. 
This highlights the model’s sensitivity to fine-grained, task-irrelevant visual cues and reveals a lack of robustness in its visual reasoning process. 
At the same time, these examples provide valuable insights into the visual factors that influence model behavior, making its weaknesses and vulnerabilities more interpretable.

\end{document}